\documentclass[12pt,a4paper]{article}

\usepackage[margin=25mm]{geometry}
\usepackage{amsmath}
\usepackage{amsfonts}
\usepackage{amssymb}
\usepackage{graphicx}
\usepackage{verbatim}
\immediate\write18{texcount -tex -sum  \jobname.tex > \jobname.wordcount.tex}
\usepackage{mathrsfs} 
\newtheorem{Theorem}{Theorem}
\usepackage{stmaryrd}
\usepackage[linesnumbered,ruled]{algorithm2e}
\usepackage{subcaption}

\usepackage{tikz}
\usetikzlibrary{positioning,chains,fit,shapes,calc}
\usepackage{multicol}
\usepackage{booktabs}

\usepackage{placeins}
\usepackage{mathtools}

\usepackage[comma,authoryear,round]{natbib}

\usepackage{chngcntr}
\counterwithin{table}{section}
\counterwithin{figure}{section}

\newcommand{\n}{\noindent}
\newcommand{\R}{\mathbb{R}} 
\newcommand{\bP}{\mathbb{P}} 
\newcommand{\bS}{\mathbb{S}} 

\newcommand{\Mf}{{\mathcal{M}}}   
\newcommand{\mcB}{{\mathcal{B}}}    							

\DeclareMathOperator*{\bigtimes}{\vartimes}
\DeclareMathOperator{\diag}{diag}
\DeclareMathOperator{\Cov}{Cov}

\DeclareMathOperator{\Deriv}{D}

\DeclareMathOperator{\Hess}{Hess}
\DeclareMathOperator{\grad}{grad}
\DeclarePairedDelimiter{\norm}{\lVert}{\rVert} 
\DeclareMathOperator{\egrad}{grad^e}

\DeclareMathOperator{\tr}{tr}

\newcommand{\la}{\langle}
\newcommand{\ra}{\rangle}

\newcommand*{\qed}{\null\nobreak\hfill\ensuremath{\square}}

\newcommand*\wc{{\mkern 2mu\cdot\mkern 2mu}}

\RequirePackage{fix-cm}
\providecommand{\keywords}[1]
{
	\small	
	\textbf{\textit{Keywords---}} Riemannian optimization, linear mixed models, nonlinear optimization, covariance estimation, REML estimation
}

\begin{document}

\title{Riemannian Optimization for Variance Estimation in Linear Mixed Models}
 
 \author{
        Lena Sembach\thanks{University of Trier,
        	FB IV - Department of Mathematics, D-54296, Trier,
        	Germany ({\tt sembach@uni-trier.de}).} 
         \and
         Jan Pablo Burgard\thanks{University of Trier,
        FB IV - Department of Economics, D-54296, Trier,
        Germany ({\tt burgardj@uni-trier.de}).}
        \and        
        Volker Schulz\thanks{University of Trier,
        FB IV - Department of Mathematics, D-54296, Trier,
        Germany ({\tt volker.schulz@uni-trier.de}).}
        }

\date{}

\maketitle

\begin{abstract}
	Variance parameter estimation in linear mixed models is a challenge for many classical nonlinear optimization algorithms due to the positive-definiteness constraint of the random effects covariance matrix. We take a completely novel view on parameter estimation in linear mixed models by exploiting the intrinsic geometry of the parameter space. We formulate the problem of residual maximum likelihood estimation as an optimization problem on a Riemannian manifold. Based on the introduced formulation, we give geometric higher-order information on the problem via the Riemannian gradient and the Riemannian Hessian. Based on that, we test our approach with Riemannian optimization algorithms numerically. Our approach yields a much better quality of the variance parameter estimates compared to existing approaches. 
\end{abstract}

\section{Introduction}
\label{intro}
Linear mixed models are a powerful tool to model linear relationships between observations if one part of the error variance can be explained by some grouping structure in the data. They can be seen as an alternative to classical linear regression models if the assumptions of independence and homoscedasticity are violated due to data being sampled from different levels of a grouping variable. Usually, they are used to analyze data from complex study designs like repeated measures, blocked or multilevel data designs and longitudinal data \citep{Gumedze2011} making them a very important class of models for many fields. Applications can be found in the field of neuroscience \citep{Gueorguieva2004}, psychology \citep{Doran2007}, statistical image analysis \citep[Chapter 12]{Demidenko2004} and many others. In its classical form, the linear mixed model is given by
\begin{align}
y = X \beta + Zb + \varepsilon.
\label{eq:lin_mixed_mod}
\end{align}
Here, the matrix $X \in \R^{n \times p}$ is the design matrix of the fixed effects with the fixed effects parameter $\beta \in \R^p$. The parameter $\beta$ is typically assumed to be fixed but unknown and needs to be estimated in applications. The matrix $Z \in \R^{n \times q}$ is the design matrix of the random effects with the corresponding random effects parameter $b \in \R^q$. The parameter $b$ is a random variable which follows a centered Gaussian distribution, i.e.
\begin{align}
b \sim \mathcal{N}(0, G)
\end{align}
for an unknown covariance matrix $G \in \R^{q \times q}$. Further, the residual error vector $\varepsilon \in \R^n$ is a random variable with independence and homoscedasticity of the single elements, that is $\varepsilon \sim \mathcal(0, \sigma^2 I_n)$, where $I_n$ denotes the identity matrix of size $n$. \\

In real-world applications, the design matrices $X$ and $Z$ as well as the responses $y$ are given. Thus, parameter estimation in the linear mixed model consists of finding estimates of the effects parameters $\beta$ and $b$ as well as the variance parameters $\sigma^2$ and $G$. For the estimation of the fixed effects parameters, a generalized least squares approach is typically used \citep{Gumedze2011}. For known variance parameters $G$ and $\sigma^2$, they are given by the following theorem. 

\begin{Theorem} \label{LME:Th:blue_blup}\cite[Lemma 1]{Gumedze2011}
	The BLUE of $\beta$ and the BLUP $b$ for known $G$ and $\sigma^2$ are given by
	\begin{align}
	\hat{\beta} &= (X^T H^{-1}X)^{-1} X^T H^{-1}y, \label{LME:eq:beta_hat} \\
	\tilde{b} &= GZ^T H^{-1} (y-X\hat{\beta}),		\label{LME:eq:b_tilde} 
	\end{align}
	where $H= ZGZ^T + I$. The covariance matrices $\Cov(\hat{\beta})$, $\Cov(\tilde{b})$ of $\hat{\beta}$, $\tilde{b}$, respectively, are given by
	\begin{align*}
	\Cov(\hat{\beta}) &= \sigma^2 (X^T H^{-1} X)^{-1}, \\
	\Cov(\tilde{b}) &= \sigma^2 GZ^T P(H) ZG,
	\end{align*}
	where 
	\begin{align}
	P(H) = H^{-1} - H^{-1} X(X^T H^{-1} X)^{-1}X^T H^{-1}.
	\label{LME:eq:P(H)}
	\end{align}
\end{Theorem}

Thus, the crucial part in parameter estimation lies in finding appropriate values for the variance parameters $\sigma^2$ and $G$. Approaches to find optimal values of the variance parameters $\sigma^2$ and $G$ are typically based on maximum likelihood estimation (\citealp{Gumedze2011}, \citealp[Section 2.1]{Demidenko2004}). The log-likelihood functions are either based on considering the marginal distribution of $y$ (\textit{ML estimation}) or based on considering the distribution of the error residuals $\tilde{\varepsilon} = y - X \hat{\beta}$ (\textit{residual or restricted maximum likelihood estimation}, REML estimation). The ML estimation is a biased estimator for the residual variance $\sigma^2$ because it does not take into account the degrees of freedom lost in estimating the parameter $\beta$ (\citealp{Gumedze2011}, \citealp[Section 2.2.5]{Pinheiro2000}). Thus, the REML estimation is typically to be preferred in applications as it results in an unbiased estimation of $\sigma^2$. It can be equally derived by marginalizing $\beta$ out of the distribution of $y$, see \cite{Bates2015_lme4}. For a thorough comparison of the two approaches, we refer to \cite{Gumedze2011}.

We focus on the REML log-likelihood in the following, but note that our novel Riemannian approach can be transferred to ML estimation in a straightforward manner. THE REML log-likelihood is given by
\citep{Gumedze2011}
\begin{align}
&l_R(\sigma^2, G) 	\notag\\
&= - \frac{1}{2} \bigg((n-p)\log(2\pi \sigma^2)  + \log\det(H) + \log\det(X^T H^{-1}X)  + \frac{(y-X\hat{\beta})^T H^{-1}(y-X\hat{\beta})}{\sigma^2} \bigg) \notag \\
&= - \frac{1}{2} \left((n-p)\log(2\pi\sigma^2) + \log\det(H) + \log\det(X^T H^{-1}X) + \frac{y^T P(H)y}{\sigma^2} \right), 
\label{LME:eq:reml}
\end{align}
where $H=ZGZ^T +I$ and 
\begin{align*}
P(H) = H^{-1} - H^{-1} X (X^T H^{-1} X)^{-1} X^T H^{-1}
\end{align*}
as in \eqref{LME:eq:P(H)}, for the derivation we refer to \citet[Section 2.2.5]{Demidenko2004} and \citet[Section 2.2.5]{Pinheiro2000}. One can show that both the log-likelihood function and the REML log-likelihood function \eqref{LME:eq:reml} are bounded from above and a maximum likelihood estimator exists under suitable conditions \citep[Theorem 4]{Demidenko2004}. More precisely, existence is ensured if the rank of the combined design matrices of fixed and random effects is less than the number of observations $n$ \citep[Section 2.17]{Demidenko2004}.\\

The objective in \eqref{LME:eq:reml} is a highly nonlinear and non-concave function which makes maximizing \eqref{LME:eq:reml} difficult in practice. Earlier works propose to use Newton-type methods \citep{Lindstrom1988} or the Expectation Maximization (EM) algorithm \citep{Dempster1981}. For this, a parameterization of the random effects covariance matrix is used, i.e., $G=G(\gamma)$, $\gamma \in \R^r$ and $r \ll q(q+1)/2$ \citep{Demidenko2004, Bates2015_lme4, Gumedze2011}, and the residual log-likelihood is optimized with respect to the vector $\gamma \in \R^r$. For the EM algorithm, the random effects parameter $b$ is considered as the latent variable and the algorithm alternates between updating the conditional expectation of $b \vert y$ and the variance parameters $\sigma^2$, $G$ \citep{Lindstrom1988}. However, the slow convergence of EM is a well-known drawback for linear mixed models (\citealp{Gumedze2011}, \citealp[Section 2.12]{Demidenko2004}). On the other hand, when starting too far from a local optimum, positive definiteness of the Hessian might not be given in Newton's method and convergence is not ensured \citep{Gumedze2011}. A modification of Newton's method for linear mixed model is the Fisher scoring algorithm, where the Hessian is replaced by the expected negative Hessian (expected information matrix) in the Newton equation \citep{Gumedze2011}. In case the linear mixed model is well-defined, the expected information matrix is positive definite, for details see \citet[Section 2.11]{Demidenko2004}. Yet, the Fisher scoring algorithm is computationally expensive \citep{Gumedze2011}. For any optimizer we need to ensure that the random effects covariance matrix $G$ is positive definite to have well-definedness in the objective \eqref{LME:eq:reml}. Typical approaches to ensure this consist in perturbing a singular iterate $G^t$ by an adjustment matrix such that we get positive definiteness \citep[Section 2.15.3]{Demidenko2004} or to use the reparameterization $G=LL^T$, where $L$ is a lower triangular matrix, via a Cholesky decomposition \cite[Section 2.15.4]{Demidenko2004}. The latter approach is used in the prominent \texttt{lme4} library \citep{Bates2015_lme4} implemented in R for minimizing the \textit{deviance} or the \textit{profiled REML} criterion. A main feature of the \texttt{lme4} library compared to other packages like the \texttt{nlme} package \citep{nlme_package, Pinheiro2000} is the possibility to model crossed random effects. Similar to some of the mentioned methods, the profiled residual log-likelihood is maximized, where $\sigma^2$ is profiled out of the log-likelihood \citep{Gumedze2011}. From an optimization perspective, this means that $\sigma^2$ is required to be optimal with respect to $G$ in every iteration. This is a strong restriction as it precludes various search directions which can result in slow convergence of optimizers, especially if the optimization algorithms start in a point that is not close to an optimum. Besides the outlined drawbacks of existing approaches, a problem in practice is that many of the aforementioned optimizers, including the \texttt{lme4}-approach, can result in singular fits, that is they hit the boundary of the feasible space and result in singular (or close to singular) $G$. This frequently occurs in practice for complex covariance structures (e.g. multiple correlated random slopes) and small to medium-sized data sets \citep{Bates2015_lme4}. Although the \texttt{lme4} package allows for singular fits, they are usually not desirable as the chances of numerical problems get higher and the optimizer of choice possibly does not converge \citep{lme4_package}. Besides, post-hoc inferential procedures may be inappropriate for singular fits \citep{lme4_package}. Another issue with models resulting in a singular fit is that the (REML) log-likelihood might not be well-defined for singular fits \citep[Section 2.15]{Demidenko2004}. \\[1ex]

In this paper, we develop a geometric algorithmic approach for variance estimation in linear mixed model. For this, we exploit the intrinsic geometry of the set of positive definite matrices by considering the manifold of positive definite matrices. This allows to use the evolving field of Riemannian optimization \citep{Absil, boumal2020intromanifolds} which deals with nonlinear optimization problems where the solution lies on a Riemannian manifold. Riemannian optimization has recently gained increasing interest in recent research due to its broad applicability to many problems in the area of engineering \citep{Lee2018}, computer vision \citep{Cherian2016}, data science \citep{Vandereycken2013} and many others \citep{Sato2021}. In the last decade, there has been a lot of research on the geometry of positive definite matrices in the field of machine learning. This is mainly motivated by a wide range of application fields raising from computer vision \citep{Cherian2016}, (medical) image analysis \citep{Horev2016, Moakher2011} to radar signal processing \citep{Arnaudon2013, Hua2017}. For the purpose of Riemannian optimization, acting on the manifold of $\bP^d$ is still quite novel. A prominent example of Riemannian optimization of $\bP^d$ is the computation of the Karcher mean \citep{Jeuris}, that is the computation of the center of given positive definite matrices. The works on fitting Gaussian mixture models with a Riemannian approach \citep{Hosseini2015,Hosseini2020,Sembach2021} showed much better results than with existing approaches. Motivated by the success of Riemannian optimization for Gaussian mixture models, we introduce a Riemannian formulation of variance estimation in linear mixed models in this work.\\ 

We derive a Riemannian formulation of REML log-likelihood maximization and based on that derive explicit expressions for the Riemannian gradient and the Riemannian Hessian. These are then used to build efficient Riemannian optimization algorithms to fit the variance parameters in linear mixed models which show promising results in terms of the estimation quality.\\[2ex]


The structure of this paper is as follows. In Chapter \ref{sec: RO}, we give an introduction to Riemannian optimization and the differential-geometric characteristics of the manifold of positive definite matrices needed for Riemannian optimization. In Chapter \ref{sec: RO_LMM}, we formulate the problem of estimating the variance parameters as a Riemannian optimization problem and derive the Riemannian gradient and Hessian for the REML objective. Based on that, we propose a Riemannian nonlinear conjugate gradient method and a Riemannian Newton trust-region method for which we show numerical results in Chapter \ref{sec: NumExp}.


\section{Riemannian Optimization}
\label{sec: RO}
To construct Riemannian optimization algorithms, we briefly state the main concepts of Optimization on manifolds or Riemannian optimization. A thorough introduction can be found in the text books \cite{Absil, boumal2020intromanifolds, Sato2021}. The concepts of Riemannian optimization are based on concepts from unconstrained Euclidean optimization algorithms and are generalized to (possibly nonlinear) manifolds. \\

\begin{figure}[h]\begin{center}
		\begin{tikzpicture}[scale=0.25]
		
		\filldraw (0,0) circle (1.35pt);
		\filldraw (14.01,-1.485) circle (1.05pt);
		\filldraw (8.195,-2.585) circle (1.15pt);
		
		\draw [line width=0.5mm, black ] (0,0) to [bend left=45] (7.8,6.3)
		to [bend left=45] (14,-1.5);
		\draw [line width=0.5mm, black ] (1.035,3) to [bend left=55] (7.1,1.7)
		to [bend left=12] (8.2,-2.6);
		\draw [line width=0.5mm, black ] (8.2,-2.57) to [bend left=25] (14,-1.47);
		\draw [line width=0.5mm, black ] (0,0) to [bend left=12] (6.1,3);
		\node at (6.5,-1) {$\Mf$};
		\filldraw (7,5) circle (4pt);				
		\node at (6,5.5) {$\theta$};

		\draw [line width=0.3mm] (4,4) -- ++(10,-2.5);
		\draw [line width=0.3mm] (4,4) -- ++(2.5,4);
		\draw [line width=0.3mm] (6.5,8) -- ++(10,-2.5);
		\draw [line width=0.3mm] (14,1.5) -- ++(2.5,4);
		\filldraw (4,4) circle (0.4pt); \filldraw (14,1.5) circle (0.4pt);
		\filldraw (6.5,8) circle (0.4pt); \filldraw (16.5,5.5) circle (0.4pt);
		\node at (18,4) {$T_{\theta} \Mf$};

		\draw [->,line width=0.5mm](7,5) -- ++(4,-1);	
		\node at (11.5,3.5) {$\xi_{\theta}$};

		\draw [dotted,line width=0.5mm] (7,5) to [bend left=25] node{.}(11,1);
		\filldraw (11,1) circle (4pt);
		\node at (11,0) {$R_{\theta}(\xi_{\theta})$};
		
		\end{tikzpicture}\caption{Retraction-based Riemannian Optimization}\label{Fig:retr_based_opt}
	\end{center}
\end{figure}
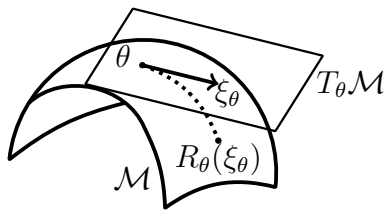

Manifolds are spaces that locally resemble vector spaces, meaning that we can locally map points on manifolds one-to-one to $\R^d$, where $d$ is the dimension of the manifold. In order to define a generalization of differentials, Riemannian optimization methods require smooth manifolds meaning that the transition mappings are smooth functions. As manifolds are in general not vector spaces, standard optimization algorithms like line-search methods cannot be directly applied as the iterates might leave the admissible set. Instead, one moves along tangent vectors in tangent spaces $T_{\theta}\Mf$, local approximations of a point $\theta \in \Mf$. A tangent space $T_{\theta}\Mf$ is a local vector space approximation around $\theta \in \Mf$ and the \textit{tangent bundle} $T\Mf$ is the disjoint union of the tangent spaces $T_{\theta}\Mf$. In Riemannian manifolds, each of the tangent spaces $T_{\theta}\Mf$, $\theta \in \Mf$ is endowed with an inner product $\langle  \cdot, \cdot \rangle_{\theta}$ that varies smoothly with $\theta$, the \textit{Riemannian metric}. The inner product is essential for Riemannian optimization methods as it admits some notion of length associated with the manifold. The optimization methods also require some local pull-back from the tangent spaces $T_{\theta} \Mf$ to the manifold $\Mf$ which can be interpreted as moving along a specific curve on $\Mf$ (dotted curve in Figure \ref{Fig:retr_based_opt}). This is realized by the concept of \textit{retractions}: Retractions are mappings from the tangent bundle $T\Mf$ to the manifold $\Mf$ with rigidity conditions: we move through the zero element $0_{\theta}$ with velocity $\xi_{\theta} \in \Mf$, i.e. $DR_{\theta} ( 0_{\theta} )[ \xi_{\theta} ] = \xi_{\theta}.$ Furthermore, the retraction of $0_{\theta} \in T_{\theta}\Mf$ at $\theta$ is $\theta$ (see Figure \ref{Fig:retr_based_opt}).


Roughly spoken, a step of a Riemannian optimization algorithm works as follows:
\begin{itemize}
	\item At iterate $\theta^t$, take a new step $\xi_{\theta^t}$ on the tangent space $T_{\theta^t}\Mf$
	\item Pull back the new step to the manifold by applying the retraction at point $\theta^t$ by setting $\theta^{t+1} = R_{\theta^t} (\xi_{\theta^t})$
\end{itemize}

Here, the crucial part that has an impact on convergence speed is updating the new iterate on the tangent space, just like in the Euclidean case. As Riemannian Optimization algorithms are a generalization of Euclidean unconstrained optimization algorithms, we thus introduce a generalization of the gradient and the Hessian.

\paragraph{Riemannian Gradient.}
In order to characterize Riemannian gradients, we need a notion of differential of functions defined on manifolds.\\

The \textit{differential} of $f : \Mf \rightarrow \R$  at $\theta$ is the linear operator
$\Deriv f(\theta) : T\Mf_{\theta} \rightarrow \R$ defined by:
\begin{align*}
\Deriv f (\theta)[v] = \frac{d}{dt} f(c(t))\biggr\rvert_{t=0},	
\end{align*}
where $c: I \rightarrow \Mf$, $0 \in I \subset \R$ is a smooth curve on $\Mf$ with $c'(0) = v$.\\

The Riemannian gradient can be uniquely characterized by the differential of the function $f$ and the inner product associated with the manifold:

The \textit{Riemannian gradient} of a smooth function $f: M \rightarrow \R$ on a Riemannian manifold is a mapping $\grad f: \Mf \rightarrow T\Mf$ such that, for all $\theta \in \Mf$, $\grad f(\theta)$ is the unique tangent vector in $T_{\theta}\Mf$ satisfying
\begin{align*}
\langle \grad f(\theta), \xi_{\theta} \rangle_{\theta} = \Deriv f(\theta) [\xi_{\theta}] 	\quad \forall \xi_{\theta} \in \Mf.
\end{align*}

\paragraph{Riemannian Hessian.} Just like we defined Riemannian gradients, we can also generalize the Hessian to its Riemannian version. To do this, we need a tool to differentiate along tangent spaces, namely the Riemannian connection (for details see \cite[Section 5.3]{Absil}).

The \textit{Riemannian Hessian} of $f: \Mf \rightarrow \R$ at $\theta$ is the linear operator $\Hess f(\theta): T_{\theta}\Mf \rightarrow T_{\theta}\Mf$ defined by
\begin{align*}
\Hess f(\theta)[\xi_{\theta}] = \nabla_{\xi_{\theta}} \grad f(\theta),
\end{align*}
where $\nabla$ is the Riemannian connection with respect to the Riemannian manifold.\\

For the generalization of some unconstrained optimization algorithms, like the L-BFGS method or the nonlinear CG method \citep{Morales2000}, it is necessary to subtract elements of different tangent spaces from each other, e.g., the gradient of the objective at two subsequent iterates. The concept of \textit{vector transport} allows to subtract tangent vectors from different tangent spaces from each other by bringing them onto the same vector space. A vector transport is a mapping $T\Mf \oplus T\Mf \rightarrow \Mf$ such that there exists a retraction $R$ with $\mathscr{T}_{\eta}(\xi) \in T_{R_{\theta}(\eta)}\Mf$, such that 
vector transport on the zero element is the identity mapping, i.e. $\mathscr{T}_{0_{\theta}}(\xi) = \xi$ is fulfilled and such that it is linear in $\xi$.

\paragraph{The Manifold of Positive Definite Matrices.}

We denote the set of (symmetric) real positive definite matrices of fixed dimension $d$ by $\bP^d$, where
\begin{align*}
\bP^{d} = \{ \Sigma \in \R^{d \times d} \vert \Sigma = \Sigma^T \text{ and } \Sigma \succ 0\}.
\end{align*}
Its tangent space is the set of symmetric real matrices of size $d$ \cite[Chapter 6]{Bhatia}, that is  for all $\Sigma \in \bP^d$, we have $T_{\Sigma} \bP^d = \bS^d$, where
\begin{align}
\bS^d = \{\xi \in \R^{d \times d} \vert \xi=\xi^T\}.
\label{eq:Cov:symm}
\end{align}

\begin{figure}[h]	\begin{center}
		\includegraphics[scale=1]{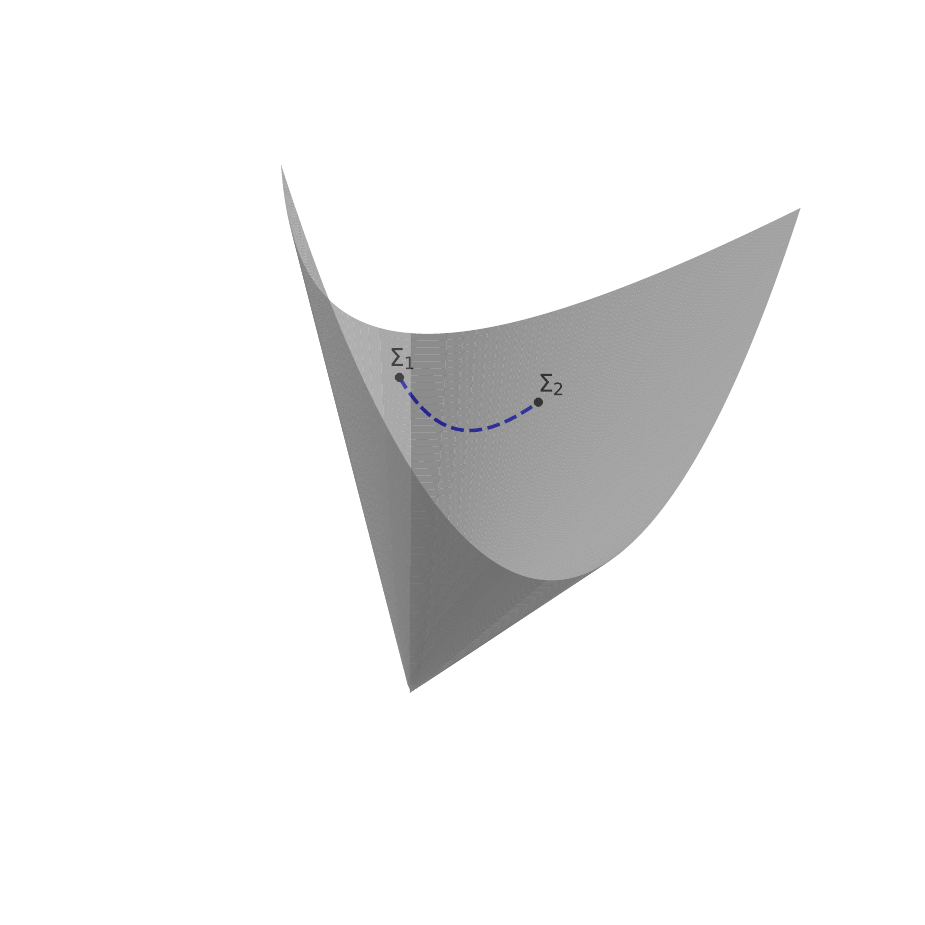}
	\end{center}
	\vspace{-2cm}
	\caption{Open cone of $\bP^2$. Visualized is the geodesics connecting $\Sigma_1, \Sigma_2 \in \bP^2$ with respect tothe affine-invariant metric  $\la \wc, \wc \ra$ (blue, dashed line). Figure inspired by \cite{Horev2016}.}\label{fig:metrics_spd}
\end{figure}

The metric which is typically associated with the manifold $\bP^d$ is the affine-invariant metric \citep{Jeuris, Bhatia} given by
\begin{align}
\la \xi, \chi \ra_{\Sigma} = \tr(\xi\Sigma^{-1} \chi \Sigma^{-1}),
 \label{eq:aff_inv_inner}
\end{align}
where $\xi, \chi \in \bS^d$ and $\Sigma \in \bP^d$. This inner product captures the intrinsic geometry of $\bP^d$ and varies smoothly with $\Sigma$. In Figure \ref{fig:metrics_spd}, the geodesic between two points $\Sigma_1, \Sigma_2 \in \bP^d$ (shortest path between the points) with respect to the Riemannian metric \eqref{eq:aff_inv_inner} is visualized which shows that the curvature of the space is captured.

Recently, the affine-invariant metric \eqref{eq:aff_inv_inner} was compared with the Bures-Wasserstein metric \citep{Han2021} for many problems relevant to machine learning. The authors of \cite{Han2021} found that the affine-invariant metric shows better results for the minimization of $f(\Sigma) = - \log\det(\Sigma)$ due to a lower condition number of the Hessian. Since this function plays a central role in the REML log-likelihood, we chose the affine-invariant metric.\\

The retraction which is commonly associated with $\bP^d$ under the Riemannian metric \eqref{eq:aff_inv_inner} is the exponential map \citep{Jeuris, Hosseini20} and given by
\begin{align}
R_{\Sigma} (\xi) = \Sigma \exp(\Sigma^{-1} \xi).					\label{retr_pd}
\end{align}
The associated vector transport is given by
\citep{Jeuris, Hosseini2015}
\begin{align}
\mathscr{T}^{\Sigma}_{\eta}(\xi) = \Sigma^{\frac{1}{2}}\exp\left(\frac{1}{2} \Sigma^{-\frac{1}{2}}\eta \Sigma^{-\frac{1}{2}}\right) \Sigma^{-\frac{1}{2}}	\eta \Sigma^{-\frac{1}{2}} 	\exp\left(\frac{1}{2} \Sigma^{-\frac{1}{2}}\eta \Sigma^{-\frac{1}{2}}\right) \Sigma^{\frac{1}{2}}.	
\label{eq:vector_transp}
\end{align}
However, usually the identity mapping is preferred for $\bP^d$ due to the high computational effort of \eqref{eq:vector_transp}. Since the set of positive definite matrices is an open set with the same tangent space at every point (although a varying inner product), the identity map can be deemed appropriate for the vector transport if step sizes do not become too large.\\

The Riemannian gradient with respect to the inner product \eqref{eq:aff_inv_inner} is given by \citep{Hosseini20, Jeuris}
\begin{align}
\grad f(\Sigma) = \Sigma \left( \grad^{sym} \bar{f}(\Sigma) \right) \Sigma = \frac{1}{2} \Sigma  \left(\grad^{e} \bar{f}(\Sigma) + \left(\grad^{e} \bar{f}(\Sigma)  \right)^T \right) \Sigma,					\label{eq:psd_grad}
\end{align}
where $\grad^{e} \bar{f}$ denotes the classical Euclidean gradient of the smooth extension $\bar{f}$ to the set of real matrices $\R^{d \times d}$ with the Euclidean metric $\la A, B \ra = \tr(A^T B)$ for $A, B \in \R^{d \times d}$.

Further, the Riemannian Hessian can be found by the Riemannian metric with respect to the inner product \eqref{eq:aff_inv_inner}. It is given by \citep{Jeuris, Sembach2021}
\begin{align}
\Hess f(\Sigma) [\xi] &= \nabla_{\xi}^{pd} \grad f(\Sigma)  = \Deriv \left(\grad f(\Sigma) \right) [\xi] - \frac{1}{2}\left(\xi \Sigma^{-1} \grad f(\Sigma) + \grad f(\Sigma) \Sigma^{-1} \xi \right),				\label{eq:RiemHess_psd}
\end{align}
where $\xi \in \bS^d$.

\section{Riemannian approach for Linear Mixed Models}
\label{sec: RO_LMM}
\label{LME:subsec:RiemAppr}
Before formulating the problem as a Riemannian optimization problem over the manifold $\bP^q$, where $G \in \bP^q$, we take a closer look at the structure of $H = I + ZGZ^T$. We note that the random effects design matrices can be decomposed into $K$ blocks, where $K$ is the number of grouping factors, i.e. $Z = (Z^{(1)}, \dots, Z^{(K)})$. Here, each of the blocks consists $M_j$ levels which are represented in the data, i.e.
\begin{align}
Z^{(j)} = \left(Z^{(j,1)}, \dots, Z^{(j, M_j)}\right),
\label{LME:eq:Zj_level}
\end{align} 
where $Z^{(j, l_j)} \in \R^{n \times q_j}$ and $q = \sum\limits_{j=1}^K M_j q_j$, see \cite{Bates2015_lme4, Gumedze2011}. The $i$-th row, $i=1,\dots, n$, of $Z^{(j)}$ denoted by $\left(Z^{(j)}\right)_i$ has possibly nonzero elements in columns $(l_j-1) q_j +1, \dots, l_j q_j$ if and only if observation $i$ arises from the $l_j$-th level of the $j$-th grouping factor. Thus, the $i$-th row of the random effects design matrix $Z$ has
\begin{align}
\sum\limits_{j=1}^K \left( M_j -1 \right)q_j
\label{LME:eq:struct_zero_Z}
\end{align}
structural zeros, i.e. the sparsity increases with the number of levels $M_j$ \citep{Bates2015_lme4}. Accordingly, the random effects parameter $b$ can be decomposed into $K$ blocks, that is $b=(b^{(1)}, \dots, b^{(K)})$ corresponding to variations introduced by $K$ different grouping factors. Each of the $b^{(j)}$ consists of $M_j$ identically and independently distributed random variables, i.e. \\ $b^{(j)} = (b^{(j,1)}, \dots, b^{(j, M_j)})$, where
\begin{align}
b^{(j, l_j)} \stackrel{iid}{\sim} \mathcal{N}(0, \Psi_j),
\label{LME:eq:b_distr}
\end{align}
where $\Psi_j \in \R^{q_j \times q_j}$. Further, we get
\begin{align}
G = \left( \begin{array}{ccc} G_1 & &  \\ & \ddots &  \\ & & G_K\end{array}\right), \qquad \text{where }  G_j = \left( \begin{array}{ccc} \Psi_j & &  \\ & \ddots &  \\ & & \Psi_j,
\end{array}\right)
\label{LME:eq:G_j_diag}
\end{align}
by a suitable ordering of the random effects. Thus, we can express the matrix $H$ in \eqref{LME:eq:reml} as

\begin{align}
H = I + ZGZ^T = I + \sum\limits_{j=1}^K Z^{(j)} G_j {Z^{(j)}}^T = I + \sum\limits_{j=1}^K \sum\limits_{l_j = 1}^{M_j} Z^{(j,l_j)} \Psi_j {Z^{(j,l_j)}}^T,
\label{LME:eq:H_decomp_Psi}
\end{align}
that is $H$ can be expressed by the unknown positive definite matrices $\Psi_j$ of the smaller dimensions $q_j$. We use this expression of the matrix $H$ for our formulation of the Riemannian optimization problem.\\

\paragraph{Riemannian formulation.} We consider the residual log-likelihood as presented in \eqref{LME:eq:reml}. The parameters of interest are the residual variance $\sigma^2$ and the random effects covariance matrix $G$ which can be expressed via the matrix $\Psi_j$ of lower dimension. For the residual variance, we introduce a variable $\eta \in \R$ and set
\begin{align*}
\eta = \log(\sigma^2).
\end{align*}
 Further, we use the derived relationship \eqref{LME:eq:H_decomp_Psi} between the covariance matrix $H$ and the matrices $\Psi_j$ and rewrite the REML log-likelihood \eqref{LME:eq:reml} as
\begin{align}
l_{R} (\theta) = - \frac{1}{2} \left((n-p)\log(2\pi)+ (n-p) \eta + \log \det(H) + \log \det (X^T H^{-1} X) + \frac{y^T P(H)y}{\exp(\eta)}\right),
\label{LME:eq:reml_riem}
\end{align}
where $\theta = (\eta, \Psi)$ for $\Psi = (\Psi_1, \dots, \Psi_K)$ and $H = I + \sum\limits_{j=1}^K \sum\limits_{l_j = 1}^{M_j} Z^{(j,l_j)} \Psi_j {Z^{(j,l_j)}}^T$. \\
The expression $P(H)$ is given by \eqref{LME:eq:P(H)}. Since the random effects covariance matrix $G$ is assumed to be positive definite, we must require that the matrices $\Psi_j \in \R^{q_j \times q_j}$ are positive definite. Thus, maximizing the REML objective \eqref{LME:eq:reml_riem} is a Riemannian optimization problem over the manifold
\begin{align}
\Mf_{LME} = \R \times \left(\bigtimes_{j=1}^K \bP^{q_j}\right), \label{LME:eq:manifold}
\end{align}
where $\bP^{q_j}$ is the manifold of positive definite matrices of dimension $q_j$ with the Riemannian metric \eqref{eq:aff_inv_inner}. The Riemannian optimization problem is summarized in the following.\\

\noindent\fbox{
	\parbox[c][12cm][s]{0.98\textwidth}{
		\vspace{0.3cm}
		\n \textbf{Optimization problem:}\\[2ex]
		Let $\theta = (\eta, \Psi)$ with $\Psi=(\Psi_1, \dots, \Psi_K)$, $\Psi_j \succ 0$, $\Psi_j \in \R^{q_j \times q_j}$. For given $X \in \R^{n \times p}$, $Z = (Z^{(1)}, \dots, Z^{(K)}) \in \R^ {n \times q}$ with $Z^{(j)} = (Z^{(j,1)}, \dots, Z^{(j, M_j)})$, $Z^{(j, l_j)} \in \R^{n \times q_j}$ and $y \in \R^n$, we consider the Riemannian optimization problem
		\begin{align}
		\min_{\theta  \in \Mf_{LME}} L_{R} (\theta) =  (n-p) \eta + \log \det(H) + \log \det (X^T H^{-1} X) + \frac{y^T P(H)y}{\exp(\eta)},	
		\label{LME:eq:riem_obj}
		\end{align}
		where 
		\begin{align*}
		H = I + \sum\limits_{j=1}^K \sum\limits_{l_j = 1}^{M_j} Z^{(j,l_j)} \Psi_j {Z^{(j,l_j)}}^T, \quad
		P(H) = H^{-1} - H^{-1} X (X^T H^{-1} X)^{-1} X^T H^{-1}
		\end{align*}
		and
		\begin{align*}
		\Mf_{LME} = \R \times \left(\bigtimes_{j=1}^K \bP^{q_j}\right).
		\end{align*}
}}\\[1.5ex]

Analogously to the profiled approach in the \texttt{lme4} package \citep{lme4_package}, we minimize minus twice the function $l_R$ in our approach, where $l_R$ is as in \eqref{LME:eq:reml}. Further, we drop the term $(n-p)\log(2\pi)$ as it does not affect the optimization.\\


\paragraph{Riemannian tools for Optimization.}
The objective function \eqref{LME:eq:riem_obj} is a smooth function over the product manifold \eqref{LME:eq:manifold} consisting of the real scalars $\R$ and the $K$ manifolds of positive definite matrices denoted by $\bP^{q_j}$. We get the tangent space  $T_{\theta}\Mf_{LME}$ of the manifold product manifold \eqref{LME:eq:manifold} as the composition of the single tangent spaces, that is
\begin{align}
T_{\theta}\Mf_{LME} = \R \times \left( \bigtimes_{j=1}^K \bS^{q_j} \right),		
\end{align}
where $\bS^{q_j}$ denotes the set of symmetric matrices of dimension $q_j$.\\

The inner product $\la \wc, \wc \ra_{\theta}$ on $T_{\theta}\Mf_{LME}$ on the product manifold can be constructed by summing the respective metrics, i.e.

\begin{align}
\langle \xi, \chi\rangle_{\theta} =\xi_{\eta}  \chi_{\eta} +  \sum\limits_{j=1}^K \tr(\Psi_j^{-1}\xi_{\Psi_j} \Psi_j^{-1} \chi_{\Psi_j}) ,
\label{LME:eq:inner_product_hs}
\end{align}
where $\theta = (\eta, \Psi) \in \Mf_{LME}$, $\Psi = (\Psi_1, \dots, \Psi_K)$, $\xi_{\theta} = (\xi_{\eta}, \xi_{\Psi}), \chi_{\theta} = (\chi_{\eta}, \chi_{\Psi}) \in T_{\theta}\Mf$ with $\xi_{\Psi} = (\xi_{\Psi_1}, \dots, \xi_{\Psi_K})$, $\chi_{\Psi} = (\chi_{\Psi_1}, \dots, \chi_{\Psi_K})$, $\xi_{\Psi_j}, \chi_{\Psi_j} \in \bS^{q_j}$, where we used \eqref{eq:aff_inv_inner}.\\

Accordingly, the retraction associated with the product manifold \eqref{LME:eq:manifold} and the inner product \eqref{LME:eq:inner_product_hs} reads 
\begin{align}
R_{\theta}(\xi_{\theta}) = \left( \begin{array}{c}\eta + \xi_{\eta} \\  \Psi_1\exp\left(\Psi_1^{-1} \xi_{\Psi_1}\right) \\  \vdots \\ \Psi_K\exp\left(\Psi_K^{-1} \xi_{\Psi_K}\right)   \end{array} \right),
\label{Retraction_probl_hs}
\end{align}
where $\theta \in \Mf_{LME}$ and $\xi_{\theta} \in T_{\theta}\Mf_{LME}$ as above, see \eqref{retr_pd}.\\

In the following, we derive expressions for the Riemannian gradient and the Riemannian Hessian of the problem \eqref{LME:eq:riem_obj}. These contribute to a deeper understanding of the underlying geometry of the optimization problem and allow for higher-order Riemannian optimizers in order to solve the optimization problem \eqref{LME:eq:riem_obj}.

\paragraph{Riemannian gradient and Hessian for Linear Mixed Models.}
We specify the Riemannian gradient for the problem \eqref{LME:eq:riem_obj} based on the inner product \eqref{LME:eq:inner_product_hs}.

\begin{Theorem} \label{LME:Th:grad} For $\theta \in \Mf_{LME}$, the Riemannian gradient $\grad L_R(\theta) \in T_{\theta}\Mf_{LME}$ of problem \eqref{LME:eq:riem_obj} is given by
	\begin{align}
	\grad L_R(\theta) = \left(\begin{array}{c}\chi_{\eta} \\ \chi_{\Psi}\end{array}\right),
	\end{align}
	where $\chi_{\Psi} = (\chi_{\Psi_1}, \dots, \chi_{\Psi_K})$ and 
	\begin{align*}
	\chi_{\eta} =  (n-p) - \frac{y^T P(H)y}{\exp(\eta)} , \qquad
	\chi_{\Psi_j}= \Psi_j \sum\limits_{l_j = 1}^{M_j}{Z^{(j,l_j)}}^T \egrad_{H}L_R(\theta) Z^{(j,l_j)} \Psi_j 
	\end{align*}
	with
	\begin{align}
	\egrad_{H} L_R(\theta) = P(H) + \frac{1}{\exp(\eta)} \egrad_{H} (y^T P(H)y),
	\label{LME:Th:egrad_H}
	\end{align}
	where 
	\begin{align}
	\egrad_{H} (y^T P(H)y) &= \bigg( - H^{-1}y y^T H^{-1} +   H^{-1} y v_1^T H^{-1} +  H^{-1} v_1 y^T H^{-1} \notag \\*
	& \qquad  \qquad - H^{-1}X (X^T H^{-1} X^T)^{-1} X^Tv_2 v_2^T X (X^T H^{-1} X^T)^{-1} X^T H^{-1} \bigg),
	\label{LME:Th:egrad_ytPy}
	\end{align}
	and $v_1 = X (X^T H^{-1} X)^{-1} X^T H^{-1} y$, $v_2 = H^{-1}y$, $P(H)$ as in \eqref{LME:eq:P(H)}.
\end{Theorem}

\n \textit{Proof.} The proof can be found in Appendix \ref{sec: Appendix}. \qed \\[2ex]

Analogously to unconstrained Euclidean optimization theory, Riemannian optimizers usually benefit from second-order information as they potentially give quadratic local convergence. Thus, besides deriving the Riemannian gradient for linear mixed models in Theorem \ref{LME:Th:grad}, we present an expression for the Riemannian Hessian of the objective \eqref{LME:eq:riem_obj} in the following.

\begin{Theorem} 
	\label{LME:Th:riemhess}
	Let $\theta \in \Mf_{LME}$ and $\xi_{\theta} \in T_{\theta}\Mf_{LME}$, $\xi_{\theta} = \left(\xi_{\eta}, \xi_{\Psi}\right)$ with $\xi_{\Psi} = \left(\xi_{\Psi_1}, \dots, \xi_{\Psi_K}\right)$.	The Riemannian Hessian of problem \eqref{LME:eq:riem_obj} is given by 
	\begin{align*}
	\Hess L_R(\theta) [\xi_{\theta}] = \left( \begin{array}{c} \zeta_{\eta} \\ \zeta_{\Psi} \end{array}\right) \in T_{\theta}\Mf_{LME},
	\end{align*}
	where
	\begin{align*}
	\zeta_{\eta} &= \frac{1}{\exp(\eta)} \bigg(\xi_{\eta} y^T P(H)y - \sum\limits_{j=1}^K y^T\bigg(h_1\left(\sum\limits_{l_j = 1}^{M_j} Z^{(j,l_j)} \xi_{\Psi_j} {Z^{(j,l_j)}}^T\right) \\*
	& \qquad \qquad \qquad  \qquad \qquad \qquad  \qquad \qquad \qquad \qquad \qquad+ h_2\left(\sum\limits_{l_j = 1}^{M_j} Z^{(j,l_j)} \xi_{\Psi_j} {Z^{(j,l_j)}}^T\right)\bigg)y  \bigg), 	
	\end{align*}
	and  $\zeta_{\Psi} = (\zeta_{\Psi_1}, \dots, \zeta_{\Psi_K}) $ with
	\begin{align*}
	\zeta_{\Psi_j} &= \Psi_j \bigg( \sum\limits_{r=1}^K \sum\limits_{l_j=1}^{M_j} {Z^{(j, l_j)}}^T\Deriv_H \left(\egrad_H L_R(\theta)\right)   [\sum\limits_{l_r=1}^{M_r} Z^{(r,l_r)} \xi_{\Psi_r} {Z^{(r,l_r)}}^T ]Z^{(j, l_j)} \\*
	& \qquad \qquad \qquad \qquad \qquad \qquad \qquad \quad - \frac{\xi_{\eta}}{\exp(\eta)}  \sum\limits_{l_j=1}^{M_j} {Z^{(j,l_j)}}^T\egrad_H(y^T P(H)y) Z^{(j,l_j)} \bigg)\Psi_j \\
	& \qquad  \qquad \qquad + \frac{1}{2} \bigg( \xi_{\Psi_j} \sum\limits_{l_j = 1}^{M_j}{Z^{(j,l_j)}}^T \egrad_{H}L_R(\theta) Z^{(j,l_j)} \Psi_j \\
	& \qquad \qquad \qquad \qquad \qquad \qquad \qquad \qquad \qquad + \Psi_j \sum\limits_{l_j = 1}^{M_j}{Z^{(j,l_j)}}^T \egrad_{H}L_R(\theta) Z^{(j,l_j)} \xi_ {\Psi_j}\bigg)
	\end{align*}
	Here,
	\begin{align*}
	\Deriv_H \left(\egrad_{H}L_R(\theta) \right) [\xi] = h_1(\xi) + h_2(\xi) + \frac{1}{\exp(\eta)} (h_3(\xi) - h_4(\xi)),
	\end{align*}
	where
	\begin{align}
	h_1(\xi) &= -H^{-1} \xi H^{-1} \label{LME:Th:Hess_h1} \\
	h_2(\xi) &=  H^{-1} \xi H^{-1} X (X^T H^{-1} X)^{-1} X^T H^{-1} + H^{-1} X (X^T H^{-1} X)^{-1} X^T H^{-1} \xi H^{-1} \notag\\
	& \qquad \qquad  -   H^{-1} X (X^T H^{-1} X)^{-1} X^T H^{-1} \xi H^{-1} X (X^T H^{-1} X)^{-1} X^T H^{-1} 
	\label{LME:Th:Hess_h2}\\
	h_3(\xi) &=   H^{-1} \xi H^{-1}y y^T H^{-1} + H^{-1} y y^T H^{-1} \xi H^{-1} \label{LME:Pr:h3}\\
	h_4(\xi) &=  -\left(h_1(\xi) + h_2(\xi)\right) y y^T H^{-1} X (X^T H^{-1} X)^{-1} H^{-1} + H^{-1}y y^T h_2(\xi)  \notag \\
	&  +   \bigg( -  \left(h_1(\xi) + h_2(\xi)\right) y y^T H^{-1} X (X^T H^{-1} X)^{-1} H^{-1} + H^{-1}y y^T h_2(\xi)\bigg)^T.
	\label{LME:Pr:h4}
	\end{align}
	The expressions $\egrad_H L_R(\theta)$, $\egrad_H(y^T P(H)y)$ are given by \eqref{LME:Th:egrad_H} and \eqref{LME:Th:egrad_ytPy}, respectively.
\end{Theorem}

\n \textit{Proof.} The proof can be found in Appendix \ref{sec: Appendix}. \qed \\[2ex]

With the Riemannian approach introduced in this paper, we formulate the objective as a function of both $\sigma^2$ and $G$, that is we estimate the residual variance and the random effects covariance matrix simultaneously. This is in contrast to maximizing the profiled log-likelihood as proposed in \cite{Bates2015_lme4} and \cite{Demidenko2004}. The approach presented here can be easily transferred to optimizing the profiled log-likelihood by substituting $\eta$ by $\hat{\eta} = \log(\hat{\sigma}(G))$, where $\hat{\sigma}(G)$ is the maximum likelihood estimator for fixed $G$.

\section{Numerical Experiments}
\label{sec: NumExp}
We tested the introduced Riemannian approach on simulated data sets. For this, we compare Riemannian optimization algorithms with the \texttt{lme4} library \citep{Bates2015_lme4}. 

\paragraph{Simulation design.}
We created data sets with $2$ grouping variables in a crossed design. We used a balanced design \cite[Section 2.2.1]{Demidenko2004}, that is we assumed that all $n$ observations are distributed equally among the $M_1$, $M_2$ levels of the two grouping factors. To generate the data sets, we followed a similar approach as suggested in \cite{Debruine2021}. We created $n=1000$ observations for our simulation design. For the fixed effects, we considered one continuous fixed effect and a fixed intercept, thus $p=2$. The fixed effects parameter was set to $\beta = (1,2)^T$ for all experiments. For the variance parameters, we set the residual variance for all experiments equal to $\sigma^2=0.1$ and created $n=1000$ realizations of a Gaussian distribution with zero mean and variance $\sigma^2 $ to get realizations of the residual error $\varepsilon$, see \eqref{eq:lin_mixed_mod}. To incorporate the random effects in the created data sets, we created $M_1$, $M_2$ categorical variables reflecting the levels of the two grouping factors. We set the number of levels for the grouping factors equal to $M_1=15$ and $M_2=10$, respectively. These were then used to build the random effects design matrix $Z$. The numerical tests were conducted for different structures in the grouping variable specific covariance matrices $\Psi_1$, $\Psi_2$ ($G_1$, $G_2$). We generated realizations of the random effects parameter ${b=(b^{(1)}, b^{(2)})}$ with a Gaussian distribution with zero mean and the covariance matrix ${G=\diag(G_1, G_2)}$ according to \eqref{LME:eq:b_distr}.\\

\n With these choices, we created the response vector $y \in \R^n$ according to the linear mixed model \eqref{eq:lin_mixed_mod}, that is
\begin{align*}
y = X \beta + Zb + \varepsilon.
\end{align*}

\n We created $100$ data sets reflecting the same linear mixed model structure with the described simulation approach. We recall that for testing different optimizers, only the variables $y, X$ and $Z$ are available and we do not know the fixed effects parameter $\beta$ or the random effects parameter $b$. 

\paragraph{Algorithmic Considerations.}
When considering the objective \eqref{LME:eq:riem_obj} for Riemannian optimization, we observe that we need to evaluate $H \in \R^{n \times n}$ in every iteration as well as terms involving its inverse $H^{-1}$. Since the number of observations $n$ is usually large in applications, precaution is required that the matrix $H$ as well as terms involving its inverse $H^{-1}$ are implemented efficiently. The matrix $H^t$ at iteration $t$, $t=0,1,2, \dots$ reads
\begin{align*}
H^t = I + Z(G^t)Z^T = I + \sum\limits_{j=1}^K Z^{(j)} (G_j)^t {Z^{(j)}}^T,
\end{align*}
where 
\begin{align*}
G_j^t = \left( \begin{array}{ccc} \Psi_j^t & &  \\ & \ddots &  \\ & & \Psi_j^t
\end{array}\right)
\end{align*}
is the random effects covariance matrix of the $j$-th grouping factor at iteration $t$ and $\Psi_j^t \in \bP^{q_j}$ is the coordinate iterate at iteration $t$, see \eqref{LME:eq:G_j_diag}. Recall that the grouping factor specific design matrices $Z^{(j)}$ are sparse. For this reason, we store the matrices ${Z^{(j)}}^T \in \R^{M_jq_j \times n}$ in a \textit{compressed sparse column (csc)} format \cite{CHOLMOD}. Further, each column of ${Z^{(j)}}^T$ consists of 
$\left( M_j -1 \right)q_j$ structural zeros, see \eqref{LME:eq:struct_zero_Z}. Thus, by construction of the matrix $G_j^t$, the positions of structural zeros in $Z^{(j)} (G_j^t)  {Z^{(j)}}^T$ are the same for all iterations $t$. This means that only the potential nonzero elements in the matrix $H$ need to be updated in every iteration which is exploited in the implementation used for the numerical results presented. Further, we need to compute factors involving the matrix inverse $({H^t})^{-1}$ in every iteration $t$. For this, we use the \textit{CHOLMOD} approach \citep{scipy, CHOLMOD} which exploits the sparsity pattern of $ZGZ^T$ for the Cholesky decomposition \citep{Bates2015_lme4}.\\

\paragraph{Choice of Optimizers.}
Equipped with the Riemannian tools and the derived formulas for the Riemannian gradient and the Hessian in Theorem \ref{LME:Th:grad} and Theorem \ref{LME:Th:riemhess}, respectively, we used higher-order Riemannian optimization algorithms to test our approach. Due to their fast local convergence close to an optimum, we applied a Riemannian Newton trust-region method (Algorithm \ref{RO:alg:RTR}) and a Riemannian nonlinear conjugate gradient method (Algorithm \ref{RO:alg:RCG}) for the problem of variance estimation in linear mixed models as presented in \eqref{LME:eq:riem_obj}. For the latter, we used the toolbox \texttt{pymanopt} \cite{Pymanopt}. For the quadratic subproblem in the Riemannian Newton trust-region algorithm (Algorithm \ref{RO:alg:RTR}), we used the truncated conjugate gradient (tCG) method \cite{Steihaug1983} as a solver. We did not use a preconditioner for the tCG method since we observed very small numbers of inner iterations in the quadratic subproblem during the conduction of the experiments. The initial trust-region radius was set to a default value at $\Delta_0=1$ and the hyperparameters of the R-NTR algorithm were set to $\rho'=0.1$, $\omega_1= 1e-3$, $\omega_2=0.99$, $\alpha_1=0.25$ and $\alpha_2 = 3.5$ according to the suggestions in \cite{Gould2005} .\\

\begin{algorithm}[h!]
	\caption{Riemannian (nonlinear) conjugate gradients method \cite[p. 182]{Absil}}
	\label{RO:alg:RCG}
	\SetAlgoLined
	\normalsize 
	\KwIn{objective $f$, vector transport $\mathscr{T}$ with associated retraction $R$, initial iterate $\theta^0 \in \Mf$}
	\KwOut{sequence of parameters $\{\theta^t\}$}
	Set $\eta^0 = - \grad f(\theta^0)$\\
	\For{$t=0,1,\dots$}{
		Compute a suitable step size $\alpha^t$ and set
		\begin{align*}
		\theta^{t+1} = R_{\theta^t}(\alpha^t \eta^t);
		\end{align*} \\
		Compute $\beta^{t+1}$ and set
		\begin{align*}
		\eta^{t+1} = - \grad f(\theta^{t+1}) + \beta^{t+1} \mathscr{T}_{\alpha^t \eta^t}(\eta^t).
		\end{align*}
	}
\end{algorithm}

\begin{algorithm}[h!]
	\caption{Riemannian trust-region method \cite[p. 142]{Absil}}
	\label{RO:alg:RTR}
	\SetAlgoLined
	\normalsize 
	\KwIn{objective $f$ with linear operator $H$, retraction $R$, initial iterate $\theta^{0} \in \Mf$, initial TR-radius $\Delta_0$, maximal TR-radius $\bar{\Delta}$, rejection threshold $\rho' \in [0, 1/4)$, acceptance parameters $0 \leq \omega_1 \le \omega_2 \leq 1, \tau_1 < 1, \tau_2 > 1$}
	\KwOut{sequence of parameters $\{\theta^t\}$}
	\For{$t=0,1,2, \dots$}{\do{
			Obtain $s^{t}$ by (approximately) solving the TR-subproblem
			\begin{align*}
			& \min_{s \in T_{\theta^{t}}\Mf} \hat{m}_{\theta^{t}}(s) =  f(\theta^{t}) + \langle \grad f(\theta^{t}), s\rangle_{\theta^{t}} + \frac{1}{2}\langle H_{t}[s],s\rangle_{\theta^{t}} \quad
			\text{ s.t. }  \langle s, s \rangle_{\theta^{t}} \leq \Delta_t^{2} ;
			\end{align*} \label{subproblem} \\
			Evaluate $\rho_t = \frac{f(\theta^{t}) - f(R_{\theta^{t}}(s^{t}))}{\hat{m}_{\theta^{t}}(0_{\theta^{t}}) - \hat{m}_{\theta^{t}}(s^{t}) }$\\
			\uIf{$\rho_t < \omega_1$}{
				$\Delta_{t+1} = \alpha_1 \Delta_t$; \label{reduce_delta}
			}\uElseIf{$\rho_t > \omega_2$ and $\norm{s^{t}}_{\theta^{t}} = \Delta_t$}{
				$\Delta_{t+1} = \min(\alpha_2\Delta_t, \bar{\Delta})$;
			}\uElse{
				$\Delta_{t+1} = \Delta_t$;
			}
			\uIf{$\rho_t > \rho'$}{
				$\theta^{t+1} = R_{\theta^{t}}(s^{t})$;
			}\uElse{
				$\theta^{t+1} = \theta^{t}$
			}
			set $t=t+1$;
	}}
\end{algorithm}

\n  Due to the popularity of the \texttt{lme4} package \citep{lme4_package}, we compared the established Riemannian optimization approach for REML estimation with the approach implemented in the \texttt{lme4} package. For this, we used the default optimizers in the \texttt{lme4} package, the BOBYQA and the Nelder Mead method \citep{Bates2015_lme4}. \\

\n We initialized the random effects covariance matrices $\Psi_j^0$ by the identity matrix and the residual variance $(\sigma^2)^0$ by its ML estimator (see \cite{Gumedze2011, Bates2011_comp}), these are the default initializations in the \texttt{lme4} package \citep{lme4_package}. We stopped all methods when either the number of iterations exceeded $1000$, when the relative difference in the objective between two subsequent iterations fell below $10^{-5}$ (for R-NTR only if we did not reject the tentative direction returned by tCG) or when the step length fell below $10^{-7}$. While the optimizers used in \texttt{lme4} are derivative-free methods, the aforementioned Riemannian optimizers compute a gradient in each iteration. Thus, we added another stopping criterion for the Riemannian optimizers, that is when the Riemannian norm of the gradient, $\norm{\grad L_R(\theta)}_{\theta}$, fell below $10^{-3}$. \\

\n All experiments were conducted in Python 3.8 on an Intel Xeon E3-1200 at 1.90 GHz with 8 cores and 16GB RAM. \\

\paragraph{Numerical Results.}
We present results for two simulation settings. The first setting is a crossed random effects design with two random intercepts which results in the dimensions $q_1 = q_2 = 1$. Thus, the structure of the grouping effect specific random effects covariance matrix is given by
\begin{align}
\Psi_j = \left(	\tau_j^2 \right),
\label{LME:eq:rand_intercept_Gj}
\end{align}
where $\tau_j$ is the standard deviation of the $j$-th random effect. The random effects variance parameters were chosen as $\tau_1 = 1.2$ and $\tau_2 = 0.9$ and the residual variance as $\sigma^2=0.1$. As outlined before, the values were initialized at  $\tau_1^0=\tau_2^0 = 1$. Table \ref{LME:tab:sim_runs_RQ1_randint} shows a summary of the optimization results of $100$ simulation runs, i.e. of $100$ generated data sets following the specified distribution. We observe that the Riemannian optimizers (R-NTR, R-CG) together with the \texttt{lme4} approach with the Nelder Mead optimizer (NELDERMEAD) show the best results in terms of the objective value $L_R$ although the deviation from the true objective (av. deviation $L_R$) is slightly higher than for the BOBYQA optimizer. In terms of the mean squared errors of the $\tau_1$, $\tau_2$ parameters (MSE $\tau_1$, $\tau_2$), we observe that both the Riemannian Newton trust-region algorithm as well as the Nelder Mead optimizer show comparable results, whereas the BOBYQA optimizer shows a higher error. In contrast, for the residual standard deviation $\sigma$, BOBYQA shows the lowest mean squared errors which is slightly below the mean squared error of the other methods. The Riemannian Newton trust-region algorithm converged in comparably few iterations, underlining the fast local convergence of the method. However, the Riemannian optimizers show much higher overall runtimes compared to the \texttt{lme4} approach with BOBYQA or Nelder Mead. This can be mainly led back to the high computational costs of evaluating the Riemannian gradient (and the Riemannian Hessian for R-NTR) in every iteration, whereas the BOBYQA and the Nelder Mead algorithms are derivative-free and have very low per-iteration costs for parameters of low dimensions (here: dimension $2$) \cite{Bates2015_lme4}.

\begin{table}[h]	
	\centering
	\footnotesize
	\caption{Simulation results for two random intercepts with ${\tau_1 = 1.2}$, ${\tau_2 = 0.9}$, ${\sigma^2=0.1}$ and $M_1 =15$, $M_2=10$.}
	\begin{tabular}{c|cccc}
		\toprule
		{} &       R-NTR &       R-CG & BOBYQA & NELDERMEAD \\
		\midrule
		av. number of iterations     &     12.01 &     55.17 &           61.43 &               36.16 \\
		av. runtime (s)  		&  6.47	 &  24.76	 &        0.05	   &            0.04 \\
		av. $L_R$      	  &  -1.5474 &  -1.5474 &        -1.4659 &            -1.5474 \\
		av. deviation $L_R$ &  2.2505 &  2.2505 &        2.0669 &            2.2505 \\
		MSE $\tau_1$      &  0.0427 	 &   0.0427  &        0.2336   &             0.0434 \\
		MSE $\tau_2$      &  0.0426 &  0.0426 &        0.1508 &            0.0425 \\
		MSE $\sigma$	  &   0.0468 &  0.0468 &        0.0446 &            0.0468 \\
		\bottomrule
	\end{tabular}
	\label{LME:tab:sim_runs_RQ1_randint}
\end{table}

The appropriateness of Riemannian optimizers for linear mixed models was tested on a second setting, where a random slope for the second grouping factor was added, i.e. $q_2=2$. The structure of the covariance matrix $G_1$ belonging to the first grouping factor $\mcB_1$ is the same as before and given by \eqref{LME:eq:rand_intercept_Gj}, whereas the structure of the covariance matrix belonging to the second grouping factor is given by

\begin{align}
\Psi_2 = \left(\begin{array}{cc} \tau_{{2}_1}^2 & \rho_2 \tau_{{2}_1} \tau_{{2}_2} \\
\rho_2 \tau_{{2}_1} \tau_{{2}_2} & \tau_{{2}_2}^2 \end{array}\right)
\label{LME:eq:rand_slope_Gj}
\end{align}

For the simulation, we set $\tau_1 = \tau_{{2}_1} = \tau_{{2}_2} = 1$ and $\rho_2=0.1$. We used the same residual variance as above, $\sigma^2=0.1$, to create $100$ data sets for the simulation. The simulation results for this setting are summarized in Table \ref{LME:tab:sim_runs_RQ1_randslope}.

\begin{table}[h]
	\caption{Simulation results for two random effects $b^{(1)} \in \R$, $b^{(2)} \in \R^2$ with $\tau_1 = \tau_{{2}_1} = \tau_{{2}_2} = 1$, $\rho_2 = 0.1$, ${\sigma^2=0.1}$ and $M_1 =15$, $M_2=10$.}
	\centering
	\footnotesize
	\begin{tabular}{c|cccc}
		\toprule
		{} &        R-NTR &        R-CG & BOBYQA & NELDERMEAD \\
		\midrule
		av. number of iterations    &  21.55 		&       30.6 	&        79.88 	&               83.28 \\
		av. runtime (s) 			&  11.79		&  		14.38 	&        0.06   &            		0.06 \\
		av. $L_R$      				&   -1.5378  	&   -1.5381 		&        -1.5305 &            -1.5393 \\
		av. deviation $L_R$ 		&   2.2922 		&   2.2933	 	&        2.2711 &            2.2969 \\
		MSE $\tau_1$      			&   0.0336 		&   0.0341 		&        0.0847 &            0.0342 \\
		MSE $\tau_{{2}_1}$    		&   0.0542 		&   0.0493 		&        0.3093 &            0.3860 \\
		MSE $\tau_{{2}_2}$    		&   0.2669 		&   0.3040 		&        0.8990 &            0.6209 \\
		MSE $\rho_2$	    		&    0.000114 &   0.000056 &        0.009555 &             0.00862 \\
		MSE $\sigma$   				&   0.0466 		&    0.0466		&        0.0464 &            0.0467 \\
		\bottomrule
	\end{tabular}
	\label{LME:tab:sim_runs_RQ1_randslope}
\end{table}
When comparing the mean squared errors for the random effects covariance matrix, we observe that the Riemannian optimizers show much better results which is visible for the second random effect in particular (MSE $\tau_{{2}_1}$, $\tau_{{2}_2}$). Those are remarkably low compared to the \texttt{lme4} optimizers. We also attain a much lower mean squared error for the correlation $\rho_2$ with the Riemannian optimizers. We observe that both Riemannian optimizers converge much faster than the \texttt{lme4} optimizers in terms of number of iterations whereas the runtimes are much higher which we also noticed for the first setting. The improved model quality of the Riemannian optimizers is remarkable as we could improve the MSE with a factor up to $8$.

\section{Conclusion}
\label{sec: Conclusion}
We took a look at variance parameter estimation from a geometric optimization perspective. For this, we introduced a novel Riemannian optimization approach for maximizing the REML log-likelihood of linear mixed models. Based on the introduced formulation, we derived explicit formulas for both the Riemannian gradient and the Riemannian Hessian which can be used to build efficient algorithms posed on Riemannian manifolds. We showed that with the Riemannian optimizers tested we received a much better quality of the parameter estimates compared to existing methods. The effect was stronger for a more complex covariance design of the random effects. These promising results give rise to further investigations for more advanced mixed models, e.g., for more complex random effects designs (e.g. more grouping factors, higher number of random slopes) or generalized mixed models that are potentially harder to fit with existing optimizers. 

\section*{Acknowledgments}
This research has been supported by the German Research Foundation (DFG) within the Research Training Group 2126:  Algorithmic Optimization, Department of Mathematics, University of Trier, Germany.

%
%

\newpage

\bibliographystyle{abbrvnat}
\bibliography{RO_LME_arxiv}

%
%

\newpage
\appendix
\section{Proofs of Theorem \ref{LME:Th:grad} and Theorem \ref{LME:Th:riemhess}}
\label{sec: Appendix}
\textbf{Proof of Theorem \ref{LME:Th:grad}}.\\
The Riemannian gradient for linear mixed models is given by the Riemannian gradient of the single components, that is 
\begin{align*}
\grad L_R(\theta) = \left(\begin{array}{c} \grad_{\eta} L_R(\theta) \\ \grad_{\Psi_1} L_R(\theta) \\ \vdots \\ \grad_{\Psi_K} L_R(\theta) \end{array}\right),
\end{align*}
where $\grad_{\eta} L_R(\theta) \in \R$, $\grad_{\Psi_j} L_R(\theta) \in \bS^{q_j}$ denotes the gradient with regard to $\eta$ and $\Psi_j$, respectively.

We first specify the Euclidean gradient $\grad_{\eta} L_R (\theta)$ of $L_R$ with respect to $\eta$. Here, the Riemannian gradient is equal to the Euclidean gradient and reads
\begin{align*}
\grad_{\eta} L_R(\theta) = (n-p) - \frac{1}{\exp(\eta)}y^T P(H)y,
\end{align*}
yielding the expression for $\chi_{\eta}$. 

The Riemannian gradient with respect to $\Psi_j$ is given by
\begin{align}
\grad_{\Psi_j} L_R(\theta) = \frac{1}{2}\Psi_j \left( \egrad_{\Psi_j} L_R(\theta) + \left(\egrad_{\Psi_j} L_R(\theta)\right)^T \right) \Psi_j,
\label{LME:Pr:grad_posdef}
\end{align}
where $\egrad_{\Psi_j} L_R(\theta)$ denotes the Euclidean gradient of the Euclidean smooth extension of $L_R(\theta)$ with respect to $\Psi_j$, see \eqref{eq:psd_grad}. We specify the Euclidean gradient $\egrad_{\Psi_j}$ for all $j=1, \dots, K$ in the following. By the chain rule, we get
\begin{align}
\egrad_{\Psi_j} L_R (\theta) = \sum\limits_{l_j=1}^{M_j} {Z^{(j,l_j)}}^T \egrad_{H}L_R(\theta) Z^{(j,l_j)},		\label{LME:Pr:egrad_Psi}
\end{align}
where $\egrad_{H} L_R (\theta)$ is the Euclidean gradient of $L_R$ with respect to the matrix $H \in \R^{n \times n}$ given by \eqref{LME:eq:H_decomp_Psi}. By again applying the chain rule, we obtain
\begin{align}
\egrad_{H}L_R(\theta) &= H^{-1} - H^{-1}X (X^T H^{-1} X^T)^{-1} X^T H^{-1}  + \frac{1}{\exp(\eta)}  \left(\egrad_H(y^T P(H)y)\right).	\label{LME:Pr:egrad_Psi2}
\end{align}
We define
\begin{align*}
u_1 (H) \coloneqq y^T H^{-1} y, \qquad
u_2 (H) \coloneqq y^T H^{-1}X (X^T H^{-1} X^T)^{-1} X^T H^{-1} y,
\end{align*}
hence the last expression in \eqref{LME:Pr:egrad_Psi2} is given by 
\begin{align}
\egrad_H(y^T P(H)y) = \egrad(u_1(H)) - \egrad(u_2(H)).
\label{LME:Pr:egrad_Psi3}
\end{align}
With the Leibniz rule, we get
\begin{align}
\egrad(u_1(H))  &= - H^{-1}y y^T H^{-1} \notag \\
\egrad(u_2(H)) &= - \bigg(H^{-1} y v_1^T H^{-1} +  H^{-1} v_1 y^T H^{-1} \notag \\
& \quad \qquad \qquad - H^{-1}X (X^T H^{-1} X^T)^{-1} X^Tv_2 v_2^T X (X^T H^{-1} X^T)^{-1} X^T H^{-1}\bigg), \label{LME:Pr:egrad_H(u)}
\end{align}
where we have set $v_1 = X (X^T H^{-1} X)^{-1} X^T H^{-1} y$ and $v_2 = H^{-1}y$.\\
We plug \eqref{LME:Pr:egrad_H(u)} into \eqref{LME:Pr:egrad_Psi3}, \eqref{LME:Pr:egrad_Psi2}, \eqref{LME:Pr:egrad_Psi} and use the relationship of the Euclidean and Riemannian gradient given by \eqref{LME:Pr:grad_posdef}. The symmetry of the Euclidean gradient $\egrad_{\Psi_j}$ with respect to $\Psi_j$ yields the expression for $\grad_{\Psi_j} L_R(\theta)$. \qed	

\vspace{1cm}

\n\textbf{Proof of Theorem \ref{LME:Th:riemhess}}.\\
The Riemannian Hessian is given by 
\begin{align}
\Hess L_{R}(\theta)[\xi_{\theta}] =  \nabla_{\theta} \grad L_{R}(\theta)	= \left(\begin{array}{c} \zeta_{\eta} \\ \left(\zeta_{\Psi_j}\right)_{j=1, \dots, K}\end{array}\right)	
= \left(\begin{array}{c} \nabla_{\xi_{\eta}	}^e \grad L_R(\theta) \\ \left(\nabla_{\xi_{\Psi_j}	}^{pd} \grad L_R(\theta) \right)_{j=1, \dots, K}	\end{array}\right)
\label{LME:eq:Proof:Hess}
\end{align}
where $\nabla^{e}_{\xi_{\eta}}$ denotes the classical Euclidean vector field differentiation along direction $\xi_{\eta}$ and $\nabla_{\xi_{\Psi_j}} ^{pd}$ denotes the Riemannian connection for positive definite matrices \cite{Jeuris}. \\

\n For the Hessian at position $\eta$ denoted by $\zeta_{\eta}$, we observe that
\begin{align*}
\zeta_{\eta} = \nabla_{\xi_{\eta}	}^e \grad L_R(\theta) = \Deriv_{\eta} \left(\egrad_{\eta}L_{R}(\theta)\right)[\xi_{\eta}]   + \sum\limits_{j=1}^K \Deriv_{\Psi_j} \left(\egrad_{\eta}L_R(\theta)\right)[\xi_{\Psi_j}].
\end{align*}
We get 
\begin{align*}
\Deriv_{\eta} \left(\egrad_{\eta}L_{R}(\theta)\right)[\xi_{\eta}] = \frac{\xi_{\eta}}{\exp(\eta)} y^T P(H) y,
\end{align*}
and
\begin{align}
\Deriv_{\Psi_j} \left(\egrad_{\eta}L_R(\theta)\right)[\xi_{\Psi_j}] &= - \frac{1}{\exp(\eta)} \Deriv_{\Psi_j} (y^T P(H)y)[\xi_{\Psi_j}] \notag \\*
&=  - \frac{1}{\exp(\eta)} \Deriv_{H} (y^T P(H)y)[\sum\limits_{l_j=1}^{M_j} Z^{(j,l_j)} \xi_{\Psi_j} {Z^{(j,l_j)}}^T].
\label{LME:eq:hess1}
\end{align}
By applying the chain rule on \eqref{LME:eq:hess1}, we get the expression for $\zeta_{\eta}$ in Theorem \ref{LME:Th:riemhess}.\\

For the Riemannian Hessian at position $\Psi_j$ denoted by $\zeta_{\Psi_j}$, using  \eqref{eq:RiemHess_psd}, we get
\begin{align}
\zeta_{\Psi_j} &= \nabla_{\xi_{\Psi_j}	}^{pd} \grad L_R(\theta)  \notag \\*
&= \Deriv_{\theta} (\grad_{\Psi_j}L_R(\theta)) [\xi_{\theta}] - \frac{1}{2} \left( \xi_{\Psi_j} {\Psi_j}^{-1} \grad_{\Psi_j} L_R(\theta) + \grad_{\Psi_j} L_R(\theta) {\Psi_j}^{-1} \xi_ {\Psi_j} \right)&& \notag \\
&= \Deriv_{\eta}(\grad_{\Psi_j} L_{R}(\theta))[\xi_{\eta}] + \sum\limits_{r=1}^K \Deriv_{\Psi_r} \left(\grad_{\Psi_j}L_R(\theta)\right)[\xi_{\Psi_r}]  \notag\\*
& \qquad \qquad \qquad \qquad \qquad  - \frac{1}{2} \left( \xi_{\Psi_j} {\Psi_j}^{-1} \grad_{\Psi_j} L_R(\theta) + \grad_{\Psi_j} L_R(\theta) {\Psi_j}^{-1} \xi_ {\Psi_j} \right) \notag \\
&=  \Deriv_{\eta}(\grad_{\Psi_j} L_{R}(\theta))[\xi_{\eta}] + \Psi_j \left( \sum\limits_{r=1}^K \Deriv_{\Psi_r} \left(\egrad_{\Psi_j}L_R(\theta)\right)[\xi_{\Psi_r}] \right) \Psi_j \notag \\*
&\qquad \qquad \qquad \qquad \qquad + \frac{1}{2} \left( \xi_{\Psi_j} \egrad_{\Psi_j} L_R(\theta) \Psi_j + \Psi_j \egrad_{\Psi_j} L_R(\theta) \xi_ {\Psi_j} \right),
\label{LME:Pr:hess_zetaPsi}
\end{align}
where we used \eqref{LME:Pr:egrad_Psi}. For the first term in \eqref{LME:Pr:hess_zetaPsi}, we get
\begin{align}
\Deriv_{\eta}(\grad_{\Psi_j} L_R(\theta))[\xi_{\eta}] = - \frac{\xi_{\eta}}{\exp(\eta)} \Psi_j \left(\sum\limits_{l_j=1}^{M_j} Z^{(j,l_j)}\egrad_H(y^T P(H)y) {Z^{(j,l_j)}}^T \right) \Psi_j,
\label{LME:Pr:hess_DfPsieta}
\end{align}
where $ \egrad_H(y^T P(H)y)$ is given by \eqref{LME:Th:egrad_ytPy}.\\

\n For the second term in \eqref{LME:Pr:hess_zetaPsi}, we get
\begin{align}
\Deriv_{\Psi_r} \left(\egrad_{\Psi_j}L_R(\theta)\right)[\xi_{\Psi_r}] &= \sum\limits_{l_j=1}^{M_j} Z^{(j, l_j)} \Deriv_{\Psi_r} \left(\egrad_H L_R(\theta)\right)[\xi_{\Psi_r}] {Z^{(j, l_j)}}^T \notag \\*
& = \sum\limits_{l_j=1}^{M_j} Z^{(j, l_j)}\Deriv_H \left(\egrad_H L_R(\theta)\right)   [\sum\limits_{l_r=1}^{M_r} Z^{(r,l_r)} \xi_{\Psi_r} {Z^{(r,l_r)}}^T ]{Z^{(j, l_j)}}^T
\label{LME:Pr:hess_DfPsi1}
\end{align}
by the chain rule. After applying the Leibniz rule on \eqref{LME:Pr:hess_DfPsi1} several times and rearrangement of terms, we get
\begin{align}
\Deriv_H \left(\egrad_{H}L_R(\theta) \right) [\xi] = h_1(\xi) + h_2(\xi) + \frac{1}{\exp(\eta)} (h_3(\xi) - h_4(\xi)),
\label{LME:Pr:hess_DFH}
\end{align}
where $h_1$, $h_2$ as in \eqref{LME:Th:Hess_h1}, \eqref{LME:Th:Hess_h2} and $h_3$, $h_4$ as in \eqref{LME:Pr:h3}, \eqref{LME:Pr:h4}.\\

\n Plugging \eqref{LME:Pr:hess_DFH} into \eqref{LME:Pr:hess_DfPsi1} and then \eqref{LME:Pr:hess_DfPsi1}, \eqref{LME:Pr:hess_DfPsieta} into \eqref{LME:Pr:hess_zetaPsi}, we get the expression for $\zeta_{\Psi_j}$ in Theorem \ref{LME:Th:riemhess}. \qed



\end{document}